\documentclass[final]{cvpr}

\usepackage{times}
\usepackage{epsfig}
\usepackage{graphicx}
\usepackage{amsmath}
\usepackage{amssymb}

\usepackage{color}
\usepackage{xcolor}

\usepackage{array} 

\usepackage[pagebackref=true,breaklinks=true,colorlinks,bookmarks=false]{hyperref}

\usepackage[caption=false]{subfig}
\usepackage{microtype}      
\usepackage[american]{babel}
\usepackage{multirow}
\usepackage{enumitem}

\def\cvprPaperID{1950} 
\def\confYear{CVPR 2021}
\begin{document}

\newcolumntype{A}{>{\centering}m{0.25\textwidth}}
\newcolumntype{B}{p{0.2\textwidth}}
\newcolumntype{C}{>{\centering\arraybackslash}m{1cm}}
\newcolumntype{D}{>{\centering\arraybackslash}m{3cm}}

\title{Contrastive Learning based Hybrid Networks for Long-Tailed Image Classification}

\author{Peng Wang$^{1}$\quad Kai Han$^{2}$\quad Xiu-Shen Wei$^{3}$\quad Lei Zhang$^{4}$ \quad Lei Wang$^{1}$ \\
$^1$University of Wollongong\quad $^2$University of Bristol \\ $^3$Nanjing University of Science and Technology \quad
$^4$Northwestern Polytechnical University\\}


\maketitle

\begin{abstract}
Learning discriminative image representations plays a vital role in long-tailed image classification because it can ease the classifier learning in imbalanced cases. Given the promising performance contrastive learning has shown recently in representation learning, in this work, we explore effective supervised contrastive learning strategies and tailor them to learn better image representations from imbalanced data in order to boost the classification accuracy thereon. Specifically, we propose a novel hybrid network structure being composed of a supervised contrastive loss to learn image representations and a cross-entropy loss to learn classifiers, where the learning is progressively transited from feature learning to the classifier learning to embody the idea that better features make better classifiers. We explore two variants of contrastive loss for feature learning, which vary in the forms but share a common idea of pulling the samples from the same class together in the normalized embedding space and pushing the samples from different classes apart. One of them is the recently proposed supervised contrastive (SC) loss, which is designed on top of the state-of-the-art unsupervised contrastive loss by incorporating positive samples from the same class. The other is a prototypical supervised contrastive (PSC) learning strategy which addresses the intensive memory consumption in standard SC loss and thus shows more promise under limited memory budget. Extensive experiments on three long-tailed classification datasets demonstrate the advantage of the proposed contrastive learning based hybrid networks in long-tailed classification.
\end{abstract}


\begin{figure}[t]
\begin{center}   
{
\includegraphics[width=\linewidth]{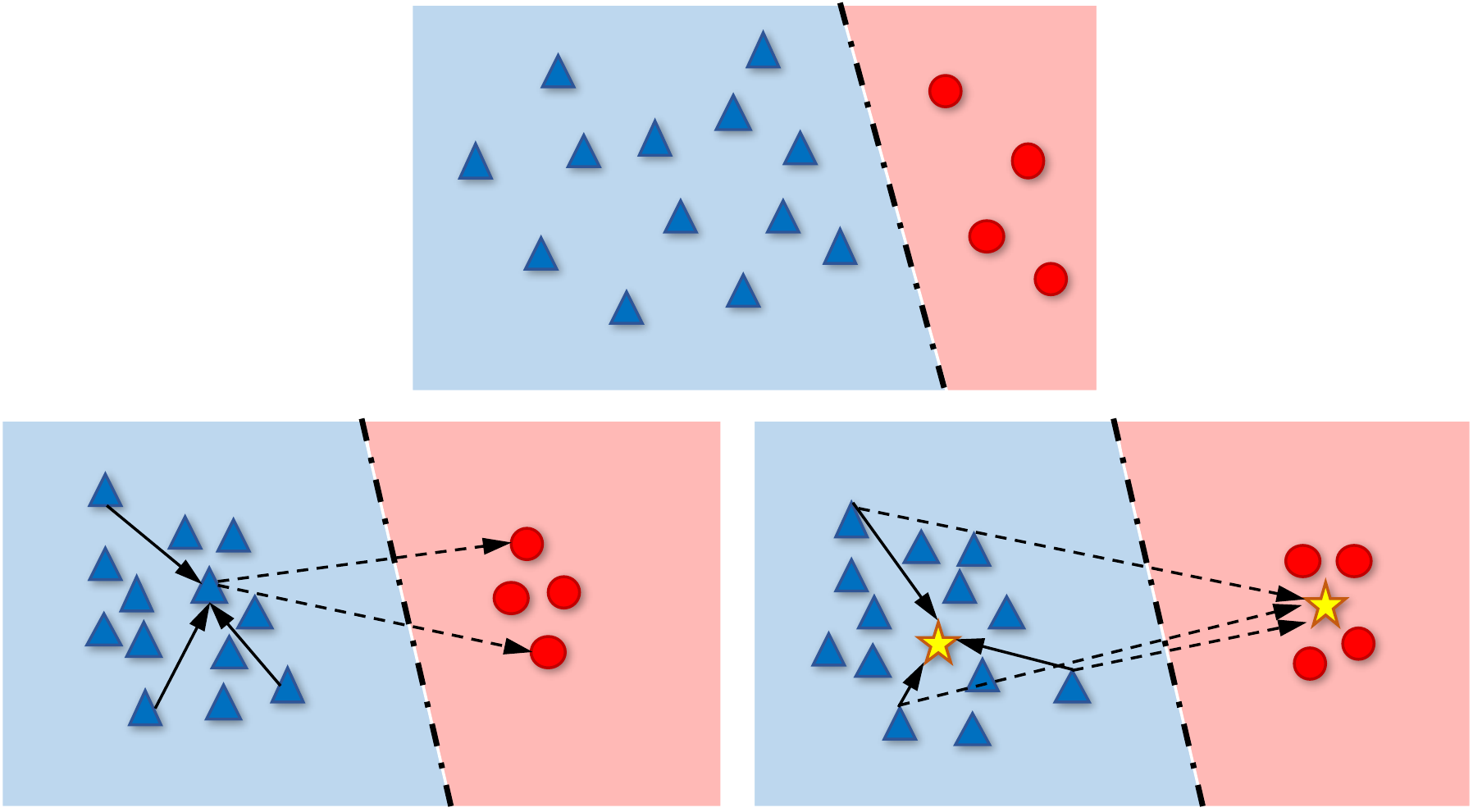}
}
\caption{Illustration of cross-entropy (upper), standard supervised contrastive (SC) (bottom left), and prototypical supervised contrastive (PSC) (bottom right) loss based feature learning for long-tailed image classification. Cross-entropy loss learns skewed features, which can result in biased classifiers. Supervised contrastive learning (bottom two) learns more intra-class compact and inter-class separable features, which ease classifier learning.  
In standard SC learning, an anchor sample together with positive samples from the same class are pulled together and the anchor is pushed away from negatives from other classes. In PSC learning, each sample is pulled towards the prototype (marked by star) of its class and pushed away from prototypes of other classes.} 
\label{fig:intro}
\end{center}
\end{figure}

\vspace*{-0.4cm}
\section{Introduction}

In the real world, the image classes are normally presented in a long-tailed distribution~\cite{openlt}. While some common classes (head classes) can have sufficient image samples, some uncommon or rare categories (tail classes) can be underrepresented by limited samples. The data imbalance poses great challenge to learning unbiased classifiers.

Most existing work addresses the data imbalance issue by mitigating the data shortage in tail classes in order to prevent the model from being dominated by the head classes. Typical methods include data re-sampling~\cite{imbalance1,imbalance2,imbalance3,SMOTE,more2016survey}, loss re-weighting~\cite{reweighting1,effnumber,ren18l2rw,equalloss}, margin modification~\cite{cao2019learning}, and data augmentation~\cite{m2m,feataug,GAMO}. Recently a new line of work is proposed which approaches long-tailed image classification by decoupling the representation learning and classifier learning into two stages~\cite{Decoupling,BBN,zhang2020balance}. The shared motivation of such work~\cite{Decoupling,BBN,zhang2020balance} is that image feature learning and classifier learning may favor different data sampling strategies and thus the focus thereon is to identify suitable sampling strategies for these two tasks. Specifically, they find under cross-entropy loss, random data sampling can benefit feature learning more while class-balanced sampling is a better option for classifier learning. 
Despite promising accuracy achieved, these methods leave the question of \emph{whether typical cross-entropy is an ideal loss for learning features from imbalanced data} untouched. Intuitively, as shown in Fig.~\ref{fig:intro}, the feature distribution learned from typical cross-entropy can be highly skewed, which can lead to biased classifiers~\cite{pmlr-v48-liud16,Huang_2016_CVPR} that harm long-tailed classification.

In this work, we explore effective contrastive learning strategies and tailor them to learn better image representations from imbalanced data in order to boost long-tailed image classification. Specifically, we propose a novel hybrid network structure composed of a contrastive loss for learning image representations and a cross-entropy loss to learn classifiers. To embody the idea that better features make better classifiers, we follow a curriculum to progressively transit the learning from feature learning to classifier learning. We realize two variants of supervised contrastive learning strategies, as shown in Fig.~\ref{fig:intro}, which vary in the forms but share a common idea of pulling the samples from the same class together in the normalized embedding space and pushing the samples from different classes apart. By doing this, less skewed features and consequently less biased classifiers are expected to be obtained. 

The first contrastive learning we explore to learn features in imbalanced scenario is the recently proposed supervised contrastive (SC) learning~\cite{SC}, which is extended from the state-of-the-art unsupervised contrastive learning~\cite{simclr} by incorporating different within-class samples as positives for each anchor. Following unsupervised contrastive learning~\cite{simclr,moco} that have two independent stages for feature learning and classifier learning, the original SC learning~\cite{SC} learns features using SC loss first and then freezes the features to learn classifiers. We argue in this paper such two-stage learning may not be an optimal choice in fully supervised scenario, which can harm the compatibility of the features and classifiers. We propose a hybrid framework to jointly learn features and classifiers, and empirically demonstrate the advantage of our joint learning mode. 

One issue of incorporating within-class positive samples in SC learning is that it leads to extra memory consumption. In SC learning~\cite{SC}, the distances to positives from the same class are contrasted with the distances to negatives from other classes, which results in memory consumption linear to the product of the positive size and negative size. 
Due to this, when under limited memory budget, the negative size needs to be shrunk. This can compromise the quality of the features learned from contrastive loss~\cite{simclr},
especially when dealing with dataset that has a large number of classes, \eg, iNaturalist~\cite{inat2017}. 

To address the aforementioned memory bottleneck from SC loss, we further propose a prototypical supervised contrastive (PSC) learning strategy, which shares the similar goal with standard SC learning but avoids explicitly sampling positives and negatives. In PSC learning, we learn a prototype for each class and force each sample to be pulled towards the prototype of its class and pushed away from prototypes of all the other classes. In this sense, the PSC strategy enables more flexible and efficient data sampling akin to softmax-based cross-entropy. It observes advantages when dealing with large-scale dataset under limited memory budget. In addition, the PSC loss has some other appealing properties that can benefit imbalanced classification, such as less sensitive to data sampling and the potential to capture finer within-class data distribution by using multiple prototypes per class.


Experiments on three long-tailed image classification datasets demonstrate the proposed contrastive learning based hybrid networks can obviously outperform the cross-entropy based counterparts and establish new state-of-the-art long-tailed image classification performance. The contributions of this work can be summarized as follows:
\begin{itemize}
\itemsep0em 
\item We propose a novel hybrid network structure for long-tailed image classification. The network is designed to be composed of a contrastive loss for feature learning and a cross-entropy loss for classifier learning. These two learning tasks are performed following a curriculum to embody the idea that better features can ease classifier learning. 
\item We explore effective supervised contrastive learning strategies to learn better features to boost long-tailed classification performance. A prototypical supervised contrastive (PSC) learning is proposed to resolve the memory bottleneck resulted from standard supervised contrastive (SC) learning. 
\item We unveil supervised contrastive learning can be a better substitute for typical cross-entropy loss for feature learning in long-tailed classification. Benefited from the better features learned, our hybrid network substantially outperforms the cross-entropy based counterparts.  
\end{itemize}

Our code is publicly available at \url{https://k-han.github.io/HybridLT}.

\section{Related Work}
Our work is closely related to both long-tailed classification and contrastive learning.

\subsection{Long-tailed image classification}

Long-tailed classification is a long-standing research problem in machine learning, where the key is to overcome the data imbalance issue~\cite{Kubat97addressingthe,nips99}. Given the great success deep neural networks have achieved in balanced classification tasks, increasing attention is being shifted to proposing neural networks based solutions for long-tailed classification. In this work, we mainly focus on the neural networks based approaches, which can be roughly divided into the following categories.


\par{\textbf{Data re-sampling}}~~
Data re-sampling is a commonly used strategy to artificially balance the imbalanced data. Two types of re-sampling techniques are under-sampling~\cite{imbalance1,more2016survey, imbalance2} and over-sampling~\cite{imbalance1, oversampling_relay, Sarafianos_2018_ECCV}. Under-sampling discards part of the data in head classes and over-sampling repetitively samples data from the tail classes. It is revealed that over-sampling can lead to overfitting to the tail classes~\cite{SMOTE, more2016survey}. Under-sampling can potentially lose information about the head classes but it may yield good results if each sample of a head class is close to other samples of the same class~\cite{more2016survey}. 

\par{\textbf{Data augmentation}}~~
As analyzed above, although over-sampling enhances the chance to see more data from the tail classes, it does not generate new information and thus leads to overfitting. One remedy is to use strong data augmentation to enrich the tail classes. Existing work approaches this goal from different angles. The work in~\cite{GAMO} uses generative model to generate new samples for tail classes as convex combination of existing instances. Another line of studies attempt to transfer the information from head classes to tail classes. In~\cite{m2m}, the authors generate data for tail classes by adding learnable noise to head samples. In another work~\cite{feataug}, the authors decompose the feature maps of images as class-generic features and class-specific features and compose new tailed data by combining class-generic features from the head image and class-specific features from a tail image. In~\cite{angular}, the intra-class angular variance is transferred from head classes to enlarge the diversity of tail classes.

\par{\textbf{Loss re-weighting}}~~
Apart from the aforementioned data-based re-balance strategies, another line of studies propose to mitigate the negative effects of data imbalance by modifying the loss functions. Loss re-weighting is one of the simple but effective ways to tailor the loss function for imbalanced classification, where the basic idea is to upweight the tailed samples and downweight the head samples in the loss function~\cite{cost-sensitive}. The existing solutions differ mainly in how to define the weights for different classes. In class-sensitive cross-entropy loss~\cite{Japkowicz02theclass}, the weight assigned to each class is inversely proportional to the number of samples. In class-balanced loss~\cite{effnumber}, the authors decide the re-weighting coefficients based on the real volumes of different classes, named effective numbers. In the work~\cite{ren18l2rw}, the weights to the training examples are optimized to minimize the loss of a held-out evaluation set.

\par{\textbf{Margin modification}}~~
It is revealed that the effect of loss re-weighting can diminish when the datasets are separable~\cite{importance_effect}. An intuitive alternative is to shift the separator closer to a dominant class~\cite{logit_adjustment}. In the work~\cite{cao2019learning}, the authors propose to integrate per-class margin into the cross-entropy loss. The margin is inversely proportional to the prior probability of a class and thus can enforce larger margins between a tail class and other classes. The work~\cite{equalloss} realizes the margin under an alternative motivation, which is to suppress the negative gradients resulted from head samples for each tailed sample.

\par{\textbf{Decoupled learning}}~~
Decoupled learning is a recent line of methods towards imbalanced classification. To identify the specific contributions of different factors to the long-tailed recognition capability, the work~\cite{Decoupling} decouples long-tailed classification into two separate stages: representation learning and classifier learning. They use cross-entropy as loss function for both of these two stages and conclude that feature learning favors random data sampling and class-balanced sampling is a better option for classifier learning. Parallel to this, the work~\cite{BBN} obtains similar conclusions empirically. In addition, a bilateral-branch network is proposed in~\cite{BBN}, where one branch uses random sampling to learn head data and the other branch uses revered sampling to emphasize tailed data. 
One common focus of these two works lies in choosing proper data sampling strategies for different learning tasks underpinning long-tailed classification. But both studies are limited to cross-entropy loss. 

\subsection{Contrastive learning}
Recently, contrastive learning has shown great promise in unsupervised representation learning~\cite{simclr,moco}. The basic idea is to learn a hidden space in which the agreement between differently augmented views of the same image is maximized by contrasting to the agreement between different images. Some key components enable the success of contrastive loss in learning useful representations include proper data augmentations, a learnable nonlinear transformation between the representation and contrastive loss, and large batch size for negative data~\cite{simclr}. Supervised contrastive (SC) learning~\cite{SC} is an extension to contrastive learning by incorporating the label information to compose positive and negative images. Following unsupervised feature learning, SC learning also adopts a two-stage learning fashion, where the first stage learns features by using contrastive loss and the second stage learns classifiers using cross-entropy loss.


\begin{figure*}[h]
\begin{center}   
{
\includegraphics[width=.9\linewidth]{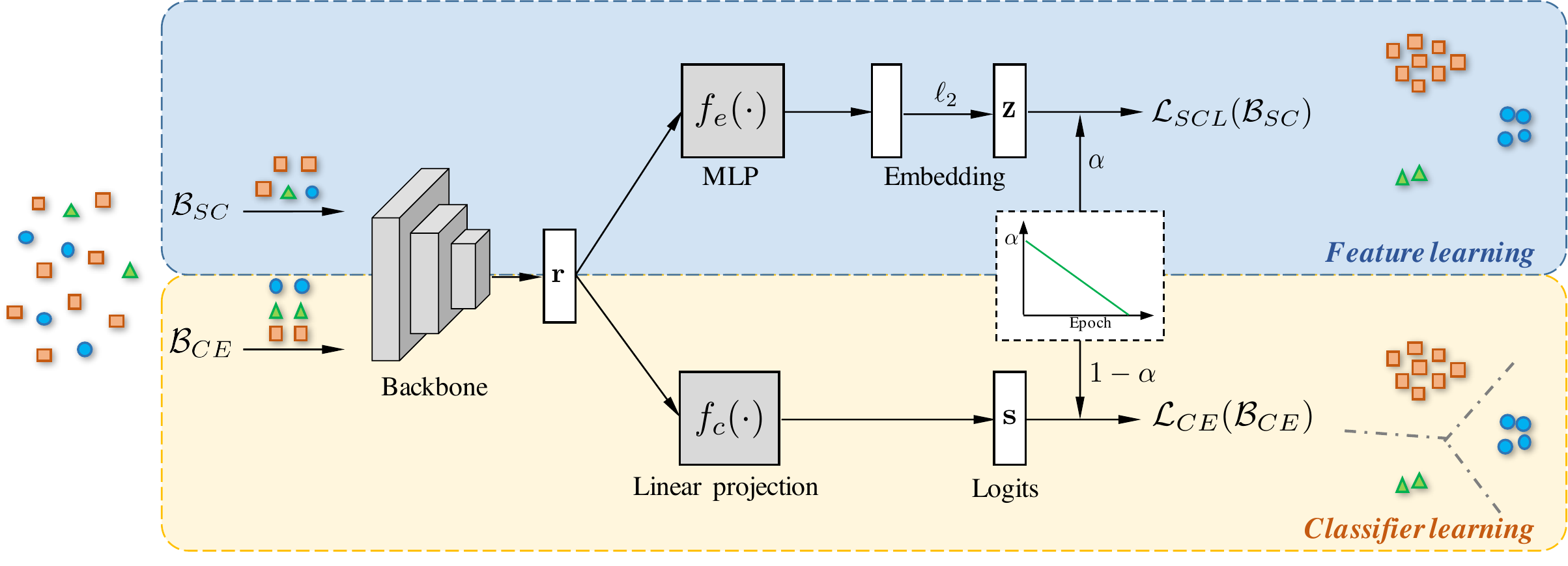}
}
\caption{Overview of the proposed contrastive learning based hybrid network structure. The network is composed of a supervised contrastive learning (SCL) based feature learning branch and a cross-entropy (CE) loss based classifier learning branch. A backbone is shared between these two branches to extract image representations, after which a non-linear MLP $f_e(\cdot)$ combined with $\ell_2$-normalization is adopted to translate the image representation for contrastive loss, and a single linear layer $f_c(\cdot)$ is applied on top of the image representation to predict classification logits. A curriculum  is designed to control the weightings of these two branches, \ie, $\alpha$ and $1-\alpha$, during network training. } 
\label{fig:framework}
\end{center}
\end{figure*}

\section{Main Approach}

In this section, we firstly present the framework for the contrastive learning based hybrid networks proposed for long-tailed classification. Then, we elaborate on the two supervised contrastive learning schemes used as part of the hybrid networks for image representation learning.

\subsection{A Hybrid Framework for Long-tailed Classification}
\label{sec:overview}
Fig.~\ref{fig:framework} shows the overview of the proposed hybrid framework for long-tailed image classification. The network consists of two branches: one contrastive learning branch for image representation learning and one cross-entropy driven branch for classifier learning. The feature learning branch aims to learn a feature space which has the property of intra-class compactness and inter-class separability. The classifier learning branch is expected to learn less biased classifier based on the discriminative features obtained from the sibling branch. To realize the idea that better features ease classifier learning and consequently result in more generalizable classifiers, we follow a curriculum~\cite{BBN} to adjust the weightings of these two branches during the training phase. Concretely, the feature learning plays a leading role at the beginning of the training, and then classifier learning gradually dominates the training.

A backbone network, \eg, ResNet~\cite{resnet}, is shared between these two learning branches to learn the image representation $\mathbf{r}\in\mathcal{R}^{D_{E}}$ for each image $\mathbf{x}$. A projection head $f_e(\cdot)$ maps the image representation $\mathbf{r}$ into a vector representation $\mathbf{z}\in\mathcal{R}^{D_S}$ which is more suitable for contrastive loss. We implement this projection head $f_e(\cdot)$ as a nonlinear multiple-layer perceptron (MLP) with one hidden layer. Such projection module is proven important in improving the representation quality of the layer before it~\cite{simclr}. Then, the $\ell_2$ normalization is applied to $\mathbf{z}$ in order that inner product can be used as distance measurements. To avoid abuse of notations, unless otherwise stated we use $\mathbf{z}$ as the normalized representation of $\mathbf{x}$ for contrastive loss computation. After that, a supervised contrastive loss $\mathcal{L}_{SCL}$ is applied on top of the normalized representations for feature learning.  The classifier learning branch is simpler which applies a single linear layer $f_c(\cdot)$ to the image representation $\mathbf{r}$ to predict the class-wise logits $\mathbf{s}\in\mathcal{R}^{D_{C}}$, which are used to compute the cross-entropy loss $\mathcal{L}_{CE}$. Due to different natures of the two loss functions, the feature learning and classifier learning branches have different data sampling strategies. The feature learning branch takes as input anchor point $\mathbf{x}_{i}$ together with positive samples $\{\mathbf{x}_i^{+}\}=\{\mathbf{x}_j|y_i=y_j,i\neq{j}\}$ from the same class and negative samples $\{\mathbf{x}_i^{-}\}=\{\mathbf{x}_j|y_j\neq{y_i}\}$ from other classes. The input batch of the feature learning branch is denoted as $\mathcal{B}_{SC} = \{\mathbf{x}_i, \{\mathbf{x}_i^{+}\}, \{\mathbf{x}_i^{-}\}\}$. The classifier learning branch directly takes image and label pairs as input $\mathcal{B}_{CE}=\{\{\mathbf{x_i},y_i\}\}$. The final loss function for the hybrid network is: 
\begin{align}
\mathcal{L}_{hybrid} = \alpha\cdot\mathcal{L}_{SCL}(\mathcal{B}_{SC}) + (1-\alpha)\cdot\mathcal{L}_{CE}(\mathcal{B}_{CE}),
\label{eq:loss}
\end{align}
where $\alpha$ is a weighting coefficient inversely proportional to the epoch number, as shown in Fig.~\ref{fig:framework}.

\subsection{Supervised contrastive loss and its memory issue}
Supervised contrastive (SC) loss~\cite{SC} is an extension to unsupervised contrastive (UC) loss~\cite{simclr}. The key difference between SC loss and UC loss lies in the composition of the positive and negative samples of an anchor image. In UC loss, the positive image is an alternatively augmented view of the anchor image. In SC loss, apart from the alternatively augmented counterpart, the positives also include some other images from the same class. In this paper, we unify all the positive images of an anchor $\mathbf{x}_i$ as $\{\mathbf{x}_i^{+}\}=\{\mathbf{x}_j|y_j=y_i, i\neq{j}\}$ (we assume different views of the same image have different indexes). The definitions for positives and negatives of $\mathbf{x}_i$ also apply to $\mathbf{z}_i$ as $\{\mathbf{z}_i^{+}\}$ and $\{\mathbf{z}_i^{-}\}$. Assuming the minibatch size is $N$, the SC loss function is written as:
\begin{align}
\mathcal{L}_{SCL}=\sum_{i=1}^{N}{\mathcal{L}_{SCL}(\mathbf{z}_i)},
\end{align}  
\begin{align}
\label{eq:sc}
\mathcal{L}_{SCL}(\mathbf{z}_i)=\frac{-1}{|\{\mathbf{z}_i^{+}\}|}\sum_{\mathbf{z}_j\in\{\mathbf{z}_i^{+}\}}{\log\frac{\exp(\mathbf{z}_i\cdot{\mathbf{z}_j}/\tau)}{\sum_{\mathbf{z}_k, k\neq{i}}{\exp(\mathbf{z}_i\cdot{\mathbf{z}_k/\tau})}}},
\end{align}
where $|\{\mathbf{z}_i^{+}\}|$ denotes the number of positive samples of anchor $\mathbf{z}_i$, and $\tau>0$ is a scalar temperature parameter.

Comparing to the UC loss~\cite{simclr}, the SC loss can flexibly incorporate arbitrary number of positives. It optimizes the agreements between such positives by contrasting against negative samples. However, the consequence of using within-class positives in SC loss is that it results in memory consumption linear to the product of positive size and negative size. For example, when one different within-class image along with an alternative view are used as positives in SC loss, the memory consumption will be doubled comparing to the UC loss with the same size of negatives. This limits the application of SC loss when under limited GPU memory budget. One solution is to shrink the size of negatives. But this can be problematic when dealing with dataset that has large number of classes because small negative size samples small fraction of negative classes, which can compromise the quality of the learned representation.  

\subsection{Prototypical supervised contrastive loss}
\label{sec:psc}
To simultaneously resolve the memory bottleneck issue and mostly retain the feature learning property of SC loss, we propose a prototypical supervised contrastive (PSC) loss. 
In PSC loss, we aim to attain similar goal of SC loss by learning a prototype for each class and force differently augmented views of each sample to be close to the prototype of their class and far away from the prototypes of the remaining classes. The benefits of using prototypes are two-fold. Firstly, it enables more flexible data sampling by avoiding explicitly sampling positives and negatives. Thus, we can flexibly adopt the data sampling strategies readily available in long-tailed classification, such as random sampling and class-balanced sampling. Secondly, data sampling efficiency is enhanced in PSC loss. In PSC loss, we contrast each sample against the prototypes of all other classes. If a dataset has $\mathcal{C}$ classes, this is essentially equivalent to a negative size of $\mathcal{C}-1$. This is practically important when dealing with dataset with large number of classes, \eg, iNaturalist~\cite{inat2017}. The PSC loss function is:
\begin{align}
\label{eq:psc}
\mathcal{L}_{PSC}(\mathbf{z}_i)=-\log\frac{\exp(\mathbf{z}_i\cdot{\mathbf{p}_{y_i}}/\tau)}{\sum_{j=1,j\neq{y_i}}^{\mathcal{C}}{\exp({\mathbf{z}_i\cdot{\mathbf{p}_{j}/\tau}})}},    
\end{align} 
where $\mathbf{p}_{y_i}$ is the prototype representation for class $y_i$, which is normalized to the unit hypersphere in $\mathcal{R}^{D_S}$ and $\mathbf{z}_i$ is the normalized representation of $\mathbf{x}_i$.

\paragraph{Extension to multiple prototypes per class}{}

In the above section, we learn one prototype per class. But PSC loss can be simply extended to multiple prototypes for each class. The rationale behind is that the samples within a class may follow a multimodal distribution, which can be better modeled by using multiple prototypes. The multiple prototype supervised contrastive (MPSC) loss function can be designed as:

\begin{align}
\mathcal{L}_{MPSC}(\mathbf{z}_i)=\frac{-1}{M}\sum_{k=1}^{M}\log\frac{w_{i,k}\exp(\mathbf{z}_i\cdot{\mathbf{p}^k_{y_i}}/\tau)}{\sum_{j=1,j\neq{y_i}}^{C}{\sum_{m=1}^{M}}\exp(\mathbf{z}_i\cdot{\mathbf{p}_j^{m}}/\tau)},
\end{align}
where $M$ is the number of prototypes per class, $\mathbf{p}^{i}_{j}$ denotes the representation for the $i$-th prototype of class $j$, and $w_{i,k} (w_{i,k} \geq 0, \sum_{k=1}^{M}w_{i,k}=1)$ denotes the affinity value between $\mathbf{z}_i$ and the $k$-th prototype of its class, which is used to control the affinity of each sample in finer level. 
We leave detailed evaluation of MPSC loss as future work.


\section{Experiments}

In this section, we firstly introduce the three long-tailed image classification datasets used for our experiments. Then we present some key implementation details of our methods. After that, we compare our proposed hybrid networks with state-of-the-art long-tailed image classification methods. Finally, some ablation studies are given
to highlight some important properties of our hybrid networks. 

\subsection{Datasets}
We conduct experiments on three long-tailed image classification datasets. Two of them, long-tailed CIFAR-10 and long-tailed CIFAR-100, are derived artificially from balanced CIFAR~\cite{cifar} datasets by re-sampling. The third dataset, iNaturalist 2018~\cite{inat2017}, is a large-scale image dataset, in which the image categories exhibit long-tailed distribution.

\par{\textbf{Long-tailed CIFAR-10 and CIFAR-100~~}} The original CIFAR-10 and CIFAR-100 datasets are balanced datasets. They consist of 50,000 training images and 10,000 validations images of size $32 \times 32$ in 10 and 100 classes respectively. Following~\cite{effnumber,cao2019learning}, the long-tailed versions are created by reducing the number of training examples per class but with the validation set unchanged. An imbalance ratio $\beta$ is used to denote the ratio between sample sizes of the most frequent and least frequent class, \ie, $\beta=N_{max}/N_{min}$. The sample size follows an exponential decay across different classes. Similar to most existing work~\cite{effnumber,cao2019learning,BBN}, we use imbalance ratios of 10, 50 and 100 in our experiments.
\par{\textbf{iNaturalist 2018~~}} The iNaturalist 2018~\cite{inat2017} is a large-scale real-world species classification dataset. It consists of 8,142 species, with 437,513 training and 24,424 validation images. The dataset observes severe imbalance in the sample sizes across different specie categories. We use the official training and validation splits for our experiments. 

\subsection{Implementation details}

In this section, we present some key implementation details for experiments on long-tailed CIFAR and iNaturalist respectively.

\noindent 
\par{\textbf{Implementation details for long-tailed CIFAR}}~
For both long-tailed CIFAR-10 and CIFAR-100, we use ResNet-32~\cite{resnet} as backbone network to extract image representation. Our hybrid network has two branches, which have independent input data as shown in Fig.~\ref{fig:framework}. The basic set of data augmentation shared by both branches include random cropping with size $32\times 32$, horizontal flip and random grayscale with probability of $0.2$. Following SC loss, we also derive different views of an image by using different data augmentations in PSC loss. In our experiments, we simply use with and without color jitter as two different augmentation views. We use batch size of $512$ for both SC and PSC based hybrid networks. The classifier learning branch uses class-wise balanced data sampling. We use SGD with a momentum of $0.9$ and weight decay of $1\times 10^{-4}$ as optimizer to train the hybrid networks. The networks are trained for 200 epochs with the learning rate being decayed by a factor of $10$ at the $120^{\text{th}}$ epoch and $160^{\text{th}}$ epoch. The initial learning rate is $0.5$. For the curriculum coefficient $\alpha$, we use a parabolic decay w.r.t the epoch number~\cite{BBN}, \ie, $\alpha=1-(T/T_{max})^2$, where $T$ denotes the current epoch number and $T_{max}$ indicates the maximum epoch number. For SC based hybrid network, the temperature $\tau$ in Eq.~\eqref{eq:sc} is fixed to be $0.1$. For PSC base hybrid network, $\tau$ is set to be $1$ for CIFAR-10 and $0.1$ for CIFAR-100.

\par{\textbf{Implementation details for iNaturalist 2018}}~
For iNaturalist 2018, following most of the existing work, we use ResNet-50~\cite{resnet} as backbone network. The data augmentation is similar to that used in long-tailed CIFAR datasets except that random cropping with size $224\times 224$ is used. To fit two NVIDIA 2080Ti GPUs, we use a batch size of 100 for both SC and PSC based hybrid networks. The networks are trained for 100 epochs using SGD with momentum 0.9 and weight decay $1\times 10^{-4}$. The initial learning rate is $0.05$, which is decayed by a factor of $10$ at epoch 60 and epoch 80. Motivated by the fact that iNaturalist has a large number of classes which can make classifier learning more difficult, we assign higher weighting to the classifier learning branch by using a linearly decayed weighting factor $\alpha$, \ie, $\alpha=1-T/T_{max}$. The temperature $\tau$ is set to be $0.1$ for both SC and PSC loss functions. For SC loss function, the number of positive samples for each anchor is fixed to $2$.

\subsection{Comparison to state-of-the-art methods}
\begin{table*}[ht]
\footnotesize
\centering
\caption{Top-1 accuracy ($\%$) on long-tailed CIFAR datasets based on ResNet-32. (Best and second best results are marked in bold.)}
\vspace{0.5em}
\label{tab:cifar}
\begin{tabular}{|D||C|C|C|C|C|C|}
\hline
Dataset           & \multicolumn{3}{c|}{Long-tailed CIFAR-10} & \multicolumn{3}{c|}{Long-tailed CIFAR-100} \\ \hline
Imbalance ratio   & 100          & 50           & 10          & 100          & 50           & 10           \\ \hline \hline
CE                & 70.36        & 74.81        & 86.39       & 38.32        & 43.85        & 55.71        \\ \hline
Focal loss~\cite{focal}        & 70.38        & 76.72        & 86.66       & 38.41        & 44.32        & 55.78        \\ \hline
CB-Focal~\cite{effnumber}          & 74.57        & 79.27        & 87.10       & 39.60        & 45.17        & 57.99        \\ \hline
CE-DRW~\cite{cao2019learning}            & 76.34        & 79.97        & 87.56       & 41.51        & 45.29        & 58.12        \\ \hline
CE-DRS~\cite{cao2019learning}            & 75.61        & 79.81        & 87.38       & 41.61        & 45.48        & 58.11        \\ \hline
LDAM-DRW~\cite{cao2019learning}          & 77.03        & 81.03        & 88.16       & 42.04        & 46.62        & 58.71        \\ \hline
CB-DA~\cite{CB-DA}             & 80.00           & 82.23        & 87.40        & 44.08        & 49.16        & 58.00           \\ \hline
M2m~\cite{m2m}               & 79.10         & --            & 87.50        & 43.50         & --            & 57.60         \\ \hline
Casual model~\cite{casual}      & {\textbf{80.6}}         & 83.60         & 88.50        & 44.10         & {\textbf{50.30}}         & 59.60         \\ \hline
BBN~\cite{BBN}               & 79.82        & 81.18        & 88.32       & 42.56        & 47.02        & 59.12        \\ \hline \hline
\textbf{Hybrid-SC} (\textbf{ours})  & {\textbf{81.40}}         & {\textbf{85.36}}        & {\textbf{91.12}}       & {\textbf{46.72}}        & {\textbf{51.87}}        & {\textbf{63.05}}        \\ \hline
\textbf{Hybrid-PSC} (\textbf{ours}) & 78.82        & {\textbf{83.86}}        & {\textbf{90.06}}       & {\textbf{44.97}}        & 48.93        & {\textbf{62.37}}        \\ \hline
\end{tabular}
\end{table*}
In this section, we compare the proposed hybrid networks, including both SC and PSC loss based networks, to existing long-tailed classification methods on long-tailed CIFAR and iNaturalist datasets, respectively. 
\par{\textbf{Experimental results on long-tailed CIFAR}} The comparison between the proposed hybrid networks and existing methods on long-tailed CIFAR datasets is presented in Table~\ref{tab:cifar}. The compared methods cover various categories of ideas for imbalanced classification, including loss re-weighting~\cite{effnumber}, margin modification~\cite{cao2019learning}, data augmentation~\cite{m2m}, decoupling~\cite{BBN} and some other newly proposed ideas~\cite{casual,CB-DA}. As can be seen from the table, our hybrid networks outperform the compared methods on almost all the settings. 

Among these methods, CE denotes the simplest baseline which directly uses cross-entropy to train the network on the long-tailed datasets. As expected, this baseline method achieves the worst performance, which reveals the limitation of cross-entropy in dealing with imbalanced data. Although the performance can be improved by using advanced loss functions tailored for long-tailed data~\cite{cao2019learning, effnumber, focal}, these methods ignore the different properties of feature learning and classifier learning. BBN~\cite{BBN} takes a step further by decoupling the head data and tailed data modeling. But several factors of BBN compromise the full potential of decoupling learning: 1)~It unifies the representation for two data streams with different properties in the penultimate layer; 2)~Cross-entropy loss is not an ideal loss for imbalanced data in both streams; 3)~The final predication in testing phase is calculated as the sum of two prediction functions from two branches with equal weights, which is inconsistent with the training phase. Our methods address such limitations in that: 1)~The projection module in our feature learning branch adapts the image representation to a space more suitable for contrastive loss; 2)~We use different loss functions to learn the features and classifiers and conclude supervised contrastive loss can be a better substitute for cross-entropy in learning features from imbalanced data; 3)~We use a single classifier learning function to predict the class labels for each sample, which avoids the gap between training and testing.  Within our methods, SC based hybrid network, a.k.a Hybrid-SC, performs better than the PSC counterpart, a.k.a Hybrid-PSC, but the latter still performs on par with or better than the compared methods.

\par{\textbf{Experimental results on iNaturalist 2018}} The experimental comparison to some existing work on iNaturalist~2018 is provided in Table~\ref{tab:inaturalist}. Again, we compare our hybrid networks to various lines of methods. Among these compared methods, Decoupling~\cite{Decoupling} and BBN~\cite{BBN} are most closely related to our proposal, which are both based on the idea of decoupled learning. The advantage of our methods over BBN has been analyzed above. On iNaturalist, Hybrid-PSC outperforms BBN by $1.8\%$. 
Classifier re-training (cRT) is a well-performed method we choose to compare in~\cite{Decoupling}. It is a two-stage method, where the first stage learns image features and the second stage freezes the features to learn the classifiers. They use cross-entropy as loss function for both stages but using different data sampling strategies. We argue this method suffers from two limitations: 1)~The two-stage learning strategy harms the compatibility between the learned features and classifiers; 2)~Cross-entropy loss is not an ideal choice for learning image features from imbalanced data.
Our hybrid network addresses the first limitation by using a curriculum based learning strategy to smoothly transit from feature learning to classifier learning. The second limitation is also observed in BBN, which can be addressed by our hybrid network as analyzed above. Our Hybrid-PSC network outperforms Decoupling~\cite{Decoupling}
by nearly $3\%$.  

Another interesting observation is that Hybrid-PSC performs better than Hybrid-SC. This result is consistent with our expectation. Note that for the two hybrid network versions, we use the same batch size of 100 for contrastive loss. This batch size is too small comparing to the number of classes in the iNaturalist dataset, which fails to provide the SC loss with sufficient negative samples to learn high-quality features~\cite{simclr}. 
PSC loss avoids this issue because, as analyzed in Sec.~\ref{sec:psc}, each sample will contrast with all the negative prototypes regardless of the batch size.  
Due to this reason, Hybrid-PSC obtains superior classification performance. Generally, we can state that the PSC based hybrid network can observe advantage over the SC loss when dealing with imbalanced dataset with large number of classes under limited GPU memory budget.

\begin{table}[h]
\footnotesize
\center
\caption{Top-1 accuracy ($\%$) on iNaturalist 2018 dataset based on ResNet-50. For Decoupling~\cite{Decoupling}, the well-performed Classifier Re-training (cRT) is reported as it is closely related to our method. By default, the methods are trained for up to 100 epochs. The number in brackets indicates the accuracy obtained by training for 200 epochs. (Best and second best results are marked in bold.)}
\vspace{0.5em}
\label{tab:inaturalist}
\begin{tabular}{|c||c|l|l|}
\hline
Dataset           & \multicolumn{3}{c|}{iNaturalist2018} \\ \hline \hline
CE                & \multicolumn{3}{c|}{57.16}           \\ \hline
CB-Focal~\cite{effnumber}          & \multicolumn{3}{c|}{61.12}           \\ \hline
CE-DRW~\cite{cao2019learning}            & \multicolumn{3}{c|}{63.73}           \\ \hline
CE-DRS~\cite{cao2019learning}            & \multicolumn{3}{c|}{63.56}           \\ \hline
LDAM-DRW~\cite{cao2019learning}          & \multicolumn{3}{c|}{\textbf{68.00}}            \\ \hline
CB-DA~\cite{CB-DA}             & \multicolumn{3}{c|}{67.55}           \\ \hline
FeatAug~\cite{feataug}           & \multicolumn{3}{c|}{65.91}           \\ \hline
Decoupling~\cite{Decoupling}        & \multicolumn{3}{c|}{65.20 (67.6)}            \\ \hline
BBN~\cite{BBN}               & \multicolumn{3}{c|}{66.29 (\textbf{69.62})}           \\ \hline \hline
\textbf{Hybrid-SC (ours)}  & \multicolumn{3}{c|}{66.74}           \\ \hline
\textbf{Hybrid-PSC (ours)} & \multicolumn{3}{c|}{\textbf{68.10} (\textbf{70.35})}            \\ \hline
\end{tabular}
\end{table}

\subsection{Ablation studies and discussions}
In this section, we conduct some ablation studies to characterize our hybrid networks. Concretely, we study whether the proposed PSC loss is less sensitive to data sampling, the advantage of using PSC loss in feature learning comparing to cross-entropy loss, and the advantage of our curriculum based joint training comparing to the two-stage learning strategy. 


\begin{table*}[t]
\footnotesize
\center
\caption{Evaluation of the sensitivity of PSC loss to data sampling. Hybrid-PSC with random PSC and Hybrid-PSC with CB-PSC denote in the PSC based hybrid network, we use random data sampling and class-balanced data sampling for the feature learning branch respectively. Classification accuracy ($\%$) on long-tailed CIFAR-100 is reported.}
\vspace{0.5em}
\label{tab:sensitivity}
\begin{tabular}{|c||l|l|l|c|l|l|c|}
\hline
Dataset                               & \multicolumn{3}{c|}{Long-tailed CIFAR-10}                                    & \multicolumn{3}{c|}{Long-tailed CIFAR-100}                                & iNaturalist 2018 \\ \hline 
Imbalance ratio                       & \multicolumn{1}{c|}{100} & \multicolumn{1}{c|}{50} & \multicolumn{1}{c|}{10} & 100                   & \multicolumn{1}{c|}{50} & \multicolumn{1}{c|}{10} & -                \\ \hline \hline
Hybrid-PSC with random PSC            & 78.82                    & 83.86                   & 90.06                   & 44.91                 & 48.93                   & 62.37                   & 68.10            \\ \hline
Hybrid-PSC with CB-PSC                & 78.84                    & 82.85                   & 89.85                   & 44.21                 & 49.66                   & 61.93                   & 67.71            \\ \hline
\end{tabular}
\end{table*}

\par{\textbf{Sensitivity of PSC loss to data sampling}}~~
In the decoupled learning work~\cite{Decoupling,BBN}, the authors find cross-entropy loss is sensitive to data sampling when it is used to learn features. Concretely, they find random sampling obviously outperforms class-wise balanced sampling for feature learning. For example, in ~\cite{Decoupling}, the class-balanced sampling can lead to around $5\%$ accuracy drop comparing to random sampling under cross-entropy loss. As PSC loss in our work has the same data sampling manner as cross-entropy loss, we verify the sensitivity of our PSC loss to data sampling in Table~\ref{tab:sensitivity}. From the table we can see, our Hybrid-PSC network achieves comparable performance by using random sampling and class-balanced sampling, which indicates our PSC can alleviate the overfitting issue resulted from over-sampling (class-balanced sampling belongs to over-sampling). We conjecture that two possible factors contribute to the insensitivity of the PSC loss on data sampling. Firstly, in PSC loss, the image features and prototypes are both $\ell_2$-normalized, which breaks the strong correlations between class frequency and feature norms. Secondly, assuming the affinity score between a sample and its prototype is $s^{y_i}_i=\mathbf{z}_i\cdot{\mathbf{p}_{y_i}}/\tau$. For a sample $\mathbf{x}_i$ with label $y_i\in\{1,2,\ldots,\mathcal{C}\}$, the gradient of the PSC loss $\mathcal{L}_{PSC}(\mathbf{z}_i)$ w.r.t $s^{y_i}_i$ is constant, and the gradient w.r.t the affinity to a prototype from a negative class $c\in\{1,2,\ldots,\mathcal{C}\}$\textbackslash ${y_i}$, 
is $\exp(s^c_i)/\sum_{y\in\{1,2,\cdots,\mathcal{C}\},y\neq{y_i}}{\exp(s^y_i)}$. The denominator excludes the dominating term of $s^{y_i}_i$ and thus results in a prominent gradient. The constant gradient for positive class and prominent gradients for negative classes can help to alleviate the overfitting in over-sampling and enhance the inter-class separability of the features.



\begin{table}[t]
\footnotesize
\center
\caption{Evaluation of the advantage of supervised contrastive losses over cross-entropy loss for feature learning in long-tailed classification. CE-CE denotes both feature learning and classifier learning adopt cross-entropy loss, \ie, our supervised contrastive loss is replaced by cross-entropy loss. Classification accuracy ($\%$) on long-tailed CIFAR-100 is reported.}
\vspace{0.5em}
\label{tab:ce-baseline}
\begin{tabular}{|c||c|l|l|}
\hline
Dataset            & \multicolumn{3}{c|}{Long-tailed CIFAR-100}                                     \\ \hline
Imbalance ratio    & 100                        & \multicolumn{1}{c|}{50} & \multicolumn{1}{c|}{10} \\ \hline \hline
CE-CE              & 41.40                       & 46.68                   & 59.14                   \\ \hline
Hybrid-SPC         & \multicolumn{1}{l|}{44.97} & 48.93                   & 62.37                   \\ \hline
Hybrid-SC & \multicolumn{1}{l|}{46.72} & 51.87                   & 63.05                   \\ \hline
\end{tabular}
\end{table}
\par{\textbf{Is PSC loss a better substitute for cross-entropy loss for feature learning?} } In this work, we claim the supervised contrastive losses are expected to learn better features from imbalanced features and consequently lead to better long-tailed classification performance. To verify this, we replace the contrastive loss in our hybrid networks with cross-entropy loss. The results are shown in Table~\ref{tab:ce-baseline}. As can be seen, when using cross-entropy to learn the image features, the performance drops significantly. 

\begin{table}[h]
\footnotesize
\center
\caption{Evaluation of the advantage of the curriculum based joint training over two-stage training. Two-stage SC denotes we train the features and classifiers in separate stages. Hybrid-SC w/o curriculum means we use equal and fixed weighting for the feature and classifier learning during the training process. Classification accuracy ($\%$) on long-tailed CIFAR-100 is reported.}
\vspace{0.5em}
\label{tab:two-stage}
\begin{tabular}{|c||c|c|c|}
\hline
Dataset                                        & \multicolumn{3}{c|}{Long-tailed CIFAR-100}                                     \\ \hline 
Imbalance ratio                                & 100                        & \multicolumn{1}{c|}{50} & \multicolumn{1}{c|}{10} \\ \hline \hline
Two-stage SC                              & 42.73                      & 46.76                   & \multicolumn{1}{c|}{60.62}  \\ \hline
\multicolumn{1}{|l||}{Hybrid-SC w/o curriculum ($\alpha=0.5$)} & \multicolumn{1}{l|}{42.58} & 47.45                   & 60.48                   \\ \hline
Hybrid-SC                                      & 46.72                      & 51.87                   & 63.05                   \\ \hline
Hybrid-SPC                                     & \multicolumn{1}{l|}{44.91} & 48.93                   & 62.37                   \\ \hline
\end{tabular}
\end{table}

\par{\textbf{Two-stage learning v.s. curriculum based joint learning}}~~ In this work, we use a curriculum to smoothly transit the training from feature learning to classifier learning. To justify the advantage of this learning strategy, we firstly choose the original two-stage SC work~\cite{SC} as our baseline, which trains the features using SC loss in the first stage and then fixes the features to train classifiers in the second stage.
From Table~\ref{tab:two-stage} we can see, this two-stage training scheme results in obviously inferior performance to our curriculum based training, because it harms the compatibility between the features and classifiers.
To further highlight the importance of the curriculum, we set the weighting coefficient $\alpha$ in Eq.~\eqref{eq:loss} to be $0.5$. Still, unsatisfactory results are obtained. When the curriculum is used, we allow the supervised contrastive losses to dominate the training first in order to fully exploit their capacity to learn discriminative features, which can benefit the classifier learning in later phase. 

\vspace*{-0.2cm}

\section{Conclusion} 

In this work, we approached long-tailed image classification by proposing a novel hybrid network, which consists of a supervised contrastive loss to learn image features and a cross-entropy loss to learn classifiers. To embody the idea that better features make better classifiers, a curriculum is followed to smoothly transit the training from feature learning to classifier learning. A new prototypical supervised contrastive loss was proposed to learn features from imbalanced data, which observes advantage under limited GPU memory budget.
Experiments on three long-tailed classification datasets showed that our proposal not only significantly outperforms existing methods but also has some other appealing properties that can benefit imbalanced classification. 
To our knowledge, this is the first work that explores how to maximize the value of supervised contrastive learning in long-tailed image classification. We will continue this direction as our future work, with the deeper exploration of MPSC as the first step. 



{\small
\bibliographystyle{ieee_fullname}
\bibliography{egbib}
}

\end{document}